\documentclass[letterpaper, 10 pt, conference]{ieeeconf}  

\IEEEoverridecommandlockouts                              

\usepackage{mathtools}
\usepackage{amsmath}
\usepackage{pdfpages}
\usepackage{cite}
\usepackage{hyperref}
\usepackage[normalem]{ulem}
\usepackage{adjustbox}
\usepackage{booktabs}
\usepackage{tabularx}
\usepackage{subcaption}
\usepackage{mathtools}
\usepackage{mathptmx}

\usepackage[ruled,vlined]{algorithm2e}

\title{\LARGE \bf
Interpreting Behaviors and Geometric Constraints as Knowledge Graphs for Robot Manipulation Control
}

\author{Chen Jiang$^{\dagger}$, Allie Wang$^{\dagger}$ and Martin Jagersand$^{\dagger}$
\thanks{$^{\dagger}$Authors are with     Department of Computing Science,
        University of Alberta, Edmonton AB., Canada, T6G 2E8.
        { 
           \tt\small \{cjiang2, luo3, mj7\}@ualberta.ca
        }
        }%
}

\begin{document}

\maketitle
\thispagestyle{empty}
\pagestyle{empty}

\begin{abstract}
In this paper, we investigate the feasibility of using knowledge graphs to interpret actions and behaviors for robot manipulation control. Equipped with an uncalibrated visual servoing controller, we propose to use robot knowledge graphs to unify behavior trees and geometric constraints, conceptualizing robot manipulation control as semantic events. The robot knowledge graphs not only preserve the advantages of behavior trees in scripting actions and behaviors, but also offer additional benefits of mapping natural interactions between concepts and events, which enable knowledgeable explanations of the manipulation contexts. Through real-world evaluations, we demonstrate the flexibility of the robot knowledge graphs to support explainable robot manipulation control.
\end{abstract}

\section{Introduction}
Interpreting motions by high-level descriptions, i.e. contexts (e.g. grasping a drawer handle versus opening the drawer), is a challenging problem for manipulation control. In practice, cost-effective control requires explainable task-level programming \cite{pedersen2016robot, sakai2022explainable}, which coordinates operations to achieve a task. A common tool to organize task-level programming is behavior trees \cite{colledanchise2018behavior}, which represent tasks as a series of behaviors and actions. However, behavior trees can overly abstract actions into high-level phrases, which do not contain the necessary low-level operation information needed for control. For example, while an action node, such as 'grasping an apple', clarifies the task goal, it does not specify the low-level motions required to achieve it.

On the other hand, understanding the manipulation contexts, which reflects human intentions, is crucial for selecting the right operations for the right tasks. For example, ViTa \cite{gridseth2016vita} constructs geometric constraints to achieve uncalibrated image-based visual servoing (UIBVS) control, where users manually associate image geometry of points and lines with specific task contexts. Further, recent advances in vision-language and foundation models \cite{levine2016end, jiang2024robot, liu2024moka, tang2025geomanip} explore the correlation between contexts, modulated as natural language, and actions, revealing geometric features as effective embodiments of affordances that are directly usable for planning low-level motions. However, programming tasks through the classic ViTa interface is not straightforward for inexperienced users, and employing foundation models \cite{liu2024moka} is computationally expensive. The question is, how can we plan tasks with image geometry, while keeping a cost-effective and intuitive understanding of the task contexts?

In summary, we wish to preserve the interpretable nature of behavior trees in scripting actions and behaviors, while establishing a clearer connection between low-level motions and high-level contexts. To this end, we propose to use knowledge graphs to achieve both goals. An example of the robot knowledge graph to relate behaviors and contexts is visualized in Figure \ref{fig:intro}. The contribution of our paper is summarized as follows:

\begin{figure}[t!]
\centering
    \includegraphics[width=\columnwidth]{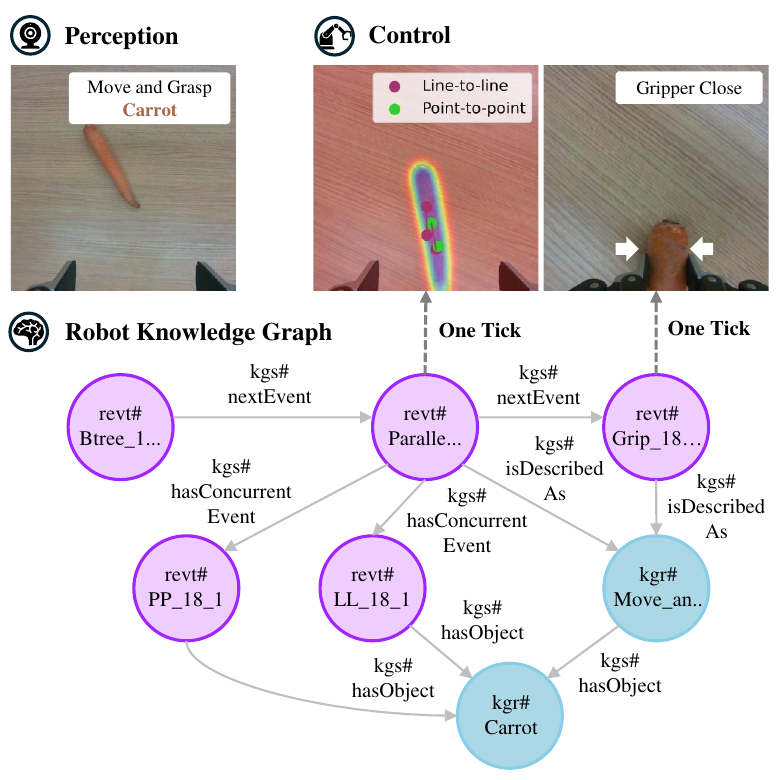}
    \caption{An example of a robot knowledge graph to conceptualize a behavior tree and geometric constraints (a point-to-point and a line-to-line constraint) for a carrot grasping task.}
     \label{fig:intro}
\end{figure}

\begin{itemize}

\item We present robot knowledge graphs, which unify behavior trees and geometric constraints to interpret actions and behaviors for uncalibrated visual servoing control.

\item Through real-world robot manipulation experiments, we demonstrate the feasibility of using robot knowledge graphs for explainable manipulation control, while also representing the robot's contextual knowledge of tasks.

\end{itemize}

\section{Related Work}
\subsection{Knowledge Graphs and Behavior Trees in Robotics}
Knowledge graphs are widely studied for their applications in knowledge representation. Systems like Ahab \cite{wang2015explicit} and RobotVQA \cite{kenfack2020robotvqa} structured concepts for visual question answering. Similar strategies to structure concepts \cite{saxena2014robobrain, jiang2020understanding, sukhwani2021dynamic, daruna2022explainable, miao2023semantic, sakib2024cooking} were applied to interpret manipulation tasks in robotics. However, graphs in those studies were used primarily for representation, without direct interaction with control.

On the other hand, behavior trees take a more direct approach through task-level programming. Recent studies \cite{cao2022behavior, dominguez2022stack, iovino2023programming, liang2024diagbt, chen2024integrating, zhou2024llm} explored the relationships between behavior trees and manipulation activities, constructing appropriate trees for control. Still, compared to knowledge graphs, the interpretability of behavior trees is inherently less. Moreover, acquiring the appropriate tree for a task is a significant challenge, where recent methods \cite{liang2024diagbt, chen2024integrating, zhou2024llm} often relied on Large-language Models (LLM) to process contexts.

\subsection{Geometric Constraints in Robotics}
Geometric constraints are manually engineered embodiments of affordance cues derived from hand-eye coordination. In robot vision, constraints could be annotated by users \cite{gridseth2016vita}, or be determined from salient vision \cite{jiang2024clipunetr, jiang2024robot, liu2024moka} to establish visuomotor control. Newer studies \cite{gao2022k, jin2022generalizable, tang2025geomanip} explored the usability of constraints as visual task functions. However, the selection of constraints was typically determined by experienced users. Automatic methods to determine constraints were rarely explored in correlation with manipulation contexts.

\section{Robot Knowledge Graphs}
Equipped with an uncalibrated visual servoing controller, we propose to extend the usage of a behavior tree as a knowledge graph to conceptualize robot manipulation control. In this section, we describe the construction of the robot knowledge graphs from the behavior trees in detail.

\subsection{Structure of Robot Knowledge Graphs}
Following Guan et al. \cite{guan2022event}, an event-centric knowledge graph is defined as $G=(V, E)$, where $V$ is the node set of entities and events. Each edge $e\in E$ is a triple $(s, p, o)$, where the subject and object $s, o\in V$, and the predicate $p$ defines the relations between entities, events, and from entities to events. For example, the sequential transition from an executable action event \texttt{A} to an action event \texttt{B} is represented as: (\texttt{A}, \texttt{nextEvent}, \texttt{B}).

\begin{figure}[h]
\centering
    \includegraphics[width=0.95\columnwidth]{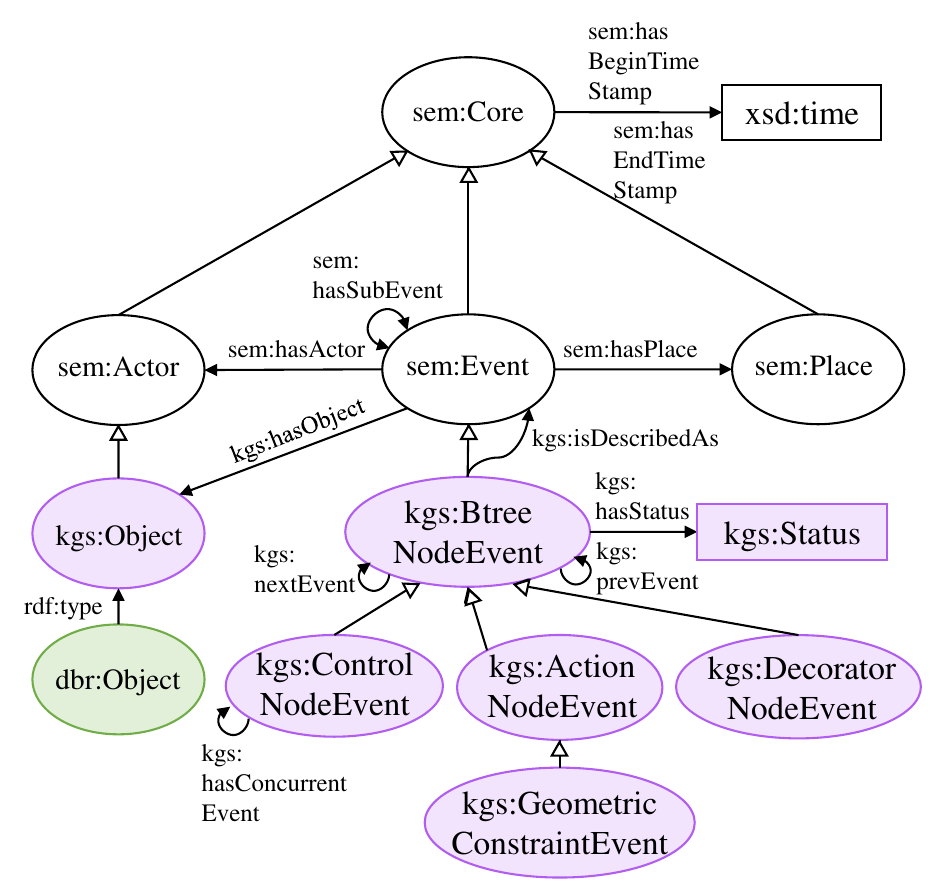}
    \caption{The schema for robot event knowledge graphs. The hollow arrow marks owl:subClassOf relation, while the regular arrow marks dedicated relations.}
     \label{fig:kg_schema}
\end{figure}

\textbf{Schema} We visualize the schematic view of the robot knowledge graphs as a UML diagram in Figure \ref{fig:kg_schema}. The design of the robot knowledge graphs follows the Simple Event Models (SEM) \cite{van2011design} to decipher an event (\texttt{sem:Event}). The event is denoted with relations to specify the timestamps (\texttt{xsd:time}), the participants (\texttt{sem:Actor}), and the location (\texttt{sem:Place}) where the event takes place. We introduce the \textbf{Knowledge Graph Schema (KGS)} namespace to structure the vocabulary of classes, properties and relations for manipulation concepts and events. For example, a manipulated object can be deciphered as \texttt{kgs:Object}, which can be disambiguated by linking to dbpedia resource (DBR), while an action node of a behavior tree can be deciphered as \texttt{kgs:BtreeNodeEvent}. The transition between events is denoted by the named relation \texttt{kgs:nextEvent}.

\textbf{Resource} Constrained by the schema, we further introduce two namespaces to host instances of entities and events: 1) \textbf{Knowledge Graph Resource (KGR)} namespace hosts the instances of manipulation concepts (e.g. \texttt{kgr:Apple\_2} is an instance of the object concept \texttt{kgs:Object}), which forms the context of a manipulation task; 2) \textbf{Robot Event (REVT)} namespace hosts the instances of robot operations and manipulation events (e.g. \texttt{revt:Grip} is an instance of the action node event \texttt{kgs:ActionNodeEvent}).

\subsection{Behavior Trees in Knowledge Graphs}
A behavior tree is a tree structure to script the flow of robot operations with control nodes and action nodes. The tree is executed by routing ticks from the root node to the leaf nodes at a given frequency, where each node returns one of $\{\textit{Success}, \textit{Failure}, \textit{Running}\}$. Our goal is to conceptualize the ticking and the routing of ticks as semantic events. Under the REVT namespace, the ticking of a ${node}_i$ is denoted as a semantic event in the form: \texttt{revt:$\mathrm{{Node}_i}$} $\xrightarrow{\mathrm{kgs:hasStatus}}$ $\mathrm{OneOf}$\{\texttt{revt:Success}, \texttt{revt:Running}, \texttt{revt:Failure}\}. The routing of a tick from ${node}_i$ to a ${node}_j$ is denoted by the transition of events as \texttt{revt:$\mathrm{{Node}_i}$} $\xrightarrow{\mathrm{kgs:nextEvent}}$ \texttt{revt:$\mathrm{{Node}_j}$}.

Furthermore, we distinguish the conceptualization among action, decorator and control nodes. An action node determines the operations or motion commands to be executed by the robot. Thus, the ticking of an action node ${node}_i$ is simply conceptualized as the event \texttt{revt:$\mathrm{{Node}_i}$}, regulated by \texttt{kgs:ActionNodeEvent} in the schema. A decorator node modifies  the return status of a single child node, which can similarly be denoted as the event \texttt{revt:$\mathrm{{Decorator}_i}$}, regulated by \texttt{kgs:DecoratorNodeEvent} in the schema.

A control node regulates the flow of operations in the behavior tree. Fundamentally, we discuss the conversion of commonly-used control nodes.

\textbf{Sequence} A \texttt{sequence} node routes the tick from its left child to the right, returning \texttt{Success} if and only if all its children nodes return \texttt{Success}. The routing of ticks is conceptualized as a transitive relation:

\begin{equation} \label{sequence}
\begin{split}
\mathrm{revt\mathord{:}{Node}_i} \xrightarrow{\mathrm{kgs:nextEvent}} \mathrm{revt\mathord{:}{Node}_j} \\
\end{split}
\end{equation}

\noindent where $\mathrm{{Node}_i}$ and $\mathrm{{Node}_j}$ are two consecutive children of the \texttt{sequence} node.

\textbf{Parallel} A \texttt{parallel} node routes ticks to all its children simultaneously, returning \texttt{Success} if and only if all children nodes return \texttt{Success}. The routing of the ticks is conceptualized as a concurrent, one-to-many relation:

\begin{equation} \label{parallel}
\begin{aligned}
\mathrm{revt\mathord{:}Parallel} &\xrightarrow{\mathrm{kgs:hasConcurrentEvent}} \mathrm{revt\mathord{:}{Node}_1}  \\
&\quad \vdots \\
&\xrightarrow{\mathrm{kgs:hasConcurrentEvent}} \mathrm{revt\mathord{:}{Node}_k}  
\end{aligned}
\end{equation}

\noindent where \{$\mathrm{{Node}_1}$, ..., $\mathrm{{Node}_k}$\} is the set of children nodes of the \texttt{parallel} node.

\textbf{Fallback} A \texttt{fallback} node routes the tick from its left child to the right, returning \texttt{Success} if any child node returns \texttt{Success} and \texttt{Failure} only if all its children return \texttt{Failure}. Similar to the \texttt{sequence} node, The routing of ticks is conceptualized as a transitive relation:

\begin{equation} \label{sequence}
\begin{split}
\mathrm{revt\mathord{:}{Node}_i} \xrightarrow{\mathrm{kgs:nextEvent}} \mathrm{revt\mathord{:}{Node}_j} , \text{where} \\
\mathrm{revt\mathord{:}{Node}_i} \xrightarrow{\mathrm{kgs:hasStatus}} \mathrm{revt\mathord{:}Failure}
\end{split}
\end{equation}

\noindent where the tick is routed from the failed $\mathrm{{Node}_i}$, to the subsequent child node $\mathrm{{Node}_j}$.

\subsection{Geometric Constraints as Behaviors}
Instead of specifying behaviors from a linguistic perspective (e.g. approach the object, sweep table, etc), we opt for a more geometric perspective in combination with an UIBVS controller. Following \cite{gridseth2016vita}, four types of geometric constraints are considered to describe the end-effector alignment given a set of points $f$:

\begin{equation} \label{visual_tasks}
\begin{aligned}
\varepsilon_{pp}(\textbf{f}) &= f_2 - f_1 \\
\varepsilon_{pl}(\textbf{f}) &= f_1 \cdot f_{34} \\
\varepsilon_{ll}(\textbf{f}) &= f_1 \cdot f_{34} + f_2 \cdot f_{34} \\
\varepsilon_{par}(\textbf{f}) &= f_{12} \times f_{34} \\
\end{aligned}
\end{equation}

\noindent where the error signals $\varepsilon_{pp}$, $\varepsilon_{pl}$, $\varepsilon_{ll}$, and $\varepsilon_{par}$ are denoted for point-to-point (p2p), point-to-line (p2l), line-to-line (l2l), and parallel-line (par) constraint, respectively. The cross product of point $f_i$ and $f_j$ computes a line $f_{ij}$. These constraints can be specified sequentially or in parallel. Given $k$ geometric constraints, we stack the error signals as follows:

\begin{equation} \label{vs_error}
\mathbf{\varepsilon} = \left( {\begin{array}{c}
    \varepsilon_1(f_1) \\
    \varepsilon_2(f_2) \\
    \vdots \\
    \varepsilon_k(f_k)) \\
  \end{array} } \right)
\end{equation}

\noindent \{$\varepsilon_1$, ..., $\varepsilon_k$\} are now optimized simultaneously, generating more complex, higher-level motion over time. The optimization is governed by the visuomotor control law:

\begin{equation} \label{vs_error}
\mathbf{\dot{\varepsilon}} = J_{u}(q)\dot{q}
\end{equation}

\noindent where $\mathbf{\dot{\varepsilon}}$ is the current visual observation of the constraint $\mathbf{\varepsilon}$, $\dot{q}$ specifies the control input of a $N$ degrees-of-freedom robot, and $J_{u}$ is the visuomotor Jacobian. 


Symbolically, we can attribute a geometric constraint as an action node of the behavior tree, regulated by \texttt{kgs:GeometricConstraintEvent} in the schema. Consequently, constraints can be simultaneously optimized, denoted by a \texttt{parallel} node:

\begin{equation} \label{parallel}
\begin{aligned}
\mathrm{revt\mathord{:}Parallel} &\xrightarrow{\mathrm{kgs:hasConcurrentEvent}} \mathrm{revt\mathord{:} {\varepsilon}_1}  \\
&\quad \vdots \\
&\xrightarrow{\mathrm{kgs:hasConcurrentEvent}} \mathrm{revt\mathord{:} {\varepsilon}_k}
\end{aligned}
\end{equation}

\noindent Similarly, constraints can be sequentially optimized, denoted by a \texttt{sequence} node: 

\begin{equation} \label{sequence}
\begin{split}
\mathrm{revt\mathord{:}{\varepsilon}_1} \xrightarrow{\mathrm{kgs:nextEvent}} \dots \xrightarrow{\mathrm{kgs:nextEvent}} \mathrm{revt\mathord{:}{\varepsilon}_k} \\
\end{split}
\end{equation}

\subsection{Control with Robot Knowledge Graphs}
With the robot knowledge graph, we present a pseudo-algorithm, in Algorithm \ref{fig:alg}, to enact real-world UIBVS control. In detail, the algorithm is summarized into three stages:

\begin{algorithm}[]
\SetAlgoLined
\textbf{Inputs:} Initial visual observation $I_{0}$. Robot knowledge base \{$G_1$, $G_2$, ...\}. Frequency $freq$. \\
\KwResult{Knowledge graph $G_i$.}
 initialize an empty $G_i$\;
 $G_i$ $\gets$ Query($I_0$, \{$G_1$, $G_2$, ...\})\;
 \While{True}{
  $events$ $\gets$ $G_i$.tick(); \\
  $G_i$ $\gets$ Conceptualize($events$); \\
  sleep($freq$); \\
 }
  \If{$G_i$.Status == $SUCCESS$}{
  Union($G_i$, \{$G_1$, $G_2$, ...\}); \
  }
 \caption{Enact robot control with the robot knowledge graph.}
 \label{fig:alg}
\end{algorithm}

\textbf{Specification} Given the real-time visual observation of the manipulation workspace as the image frame $I_0$, first, the robot analyzes the task context, which can either be specified by a user, or be determined from $I_0$. Then, the robot queries a knowledge base, which is a union of graphs \{$G_1$, $G_2$, ...\}, where each graph represents a successful manipulation experience, and outputs the set of behaviors and geometric constraints to perform the task.

\textbf{Execution} With the determined behaviors hosted inside a knowledge graph $G_i$, the robot ticks $G_i$ at a fixed frequency $freq$ (1Hz for all tasks), performing UIBVS control and completing the manipulation task. In the process, the ticking and the routing of ticks are recorded, conceptualizing the temporal transitions of behaviors as events in $G_i$.

\textbf{Abstraction} When the task is successfully completed, $G_i$ is updated into the robot knowledge base, further enriching the robot's common senses over a series of manipulation tasks. However, instantiating every tick leads to redundant instances of events unnecessarily populating $G_i$. As a post-processing step, we collapse instances of events into more abstract views before the updating step. For example, the abstraction of ticking $node_i$ is denoted as: \texttt{revt:$\mathrm{{Node}_i}$} $\xrightarrow{\mathrm{sem:hasSubEvent}}$ \texttt{revt:$\mathrm{{Node}^t_i}$}, where \texttt{revt:$\mathrm{{Node}^t_i}$} denotes the instance of the abstract event \texttt{revt:$\mathrm{{Node}_i}$}, captured when $node_i$ is ticked at timestamp $t$. 




\section{Experimental Evaluations}
\subsection{Evaluation Settings}
\textbf{Correctness of Constraints} We evaluate the explainability of knowledge graphs to determine geometric constraints from contexts. Three methods are compared:

\begin{itemize}

    \item \textbf{Symbolic Method:} Following the proposed Algorithm \ref{fig:alg}, the symbolic method takes natural language phrases (e.g., "grasp apple") as inputs, queries a knowledge base of past successful manipulation tasks, and outputs the most relevant geometric constraints for the context. 

    \item \textbf{CLIP-based Method:} The CLIP method constructs a zero-shot classifier using CLIP image encoder. First, initial frames of past successful manipulation tasks are encoded as reference embeddings. Then, the current visual observation is encoded. Cosine similarity is computed between the current and the reference embeddings. The geometric constraints from the most similar past task are selected as the output.

    \item \textbf{GPT-4o Method:} Similar to \cite{jiang2024robot}, prompted with the initial visual observation of the workspace and natural language phrases describing the goal of the task, GPT-4o infers the most relevant selection of geometric constraints.
    
\end{itemize}

\noindent From \cite{jiang2024clipunetr}, 19 demonstrations of moving and grasping tasks are annotated, forming a knowledge base. The annotated knowledge base is used as the expert knowledge base for the symbolic, and CLIP-based methods. Evaluation is conducted on 12 demonstrations. 

\textbf{Concurrent Constraints} We study the viability of scripting geometric constraints sequentially vs. concurrently. 5 trials are conducted for each of two tasks: 1) grasping a marker pen (with $\varepsilon_{par}$ and $\varepsilon_{p2p}$); and 2) grasping a drink can (with $\varepsilon_{p2l}$ and $\varepsilon_{p2p}$). Velocity control is employed across all trials. The average success rate and time are reported. 


\textbf{Real-world Robot Control} We investigate the applicability of robot knowledge graphs by performing two categories of real-world manipulation tasks using everyday objects: 1) Pick-and-place, and 2) Composition. Pick-and-place tasks generally require one or two pairs of constraints to complete the task, while composition tasks are more complex, requiring more than four behaviors or constraints. In addition, both waypoint and velocity control modes are evaluated. Waypoint control moves the end effector incrementally between waypoints as blocking actions, while velocity control continuously adjusts the end effector's position in real time as non-blocking actions. Each control mode is allowed 3 attempts. A mode is considered successful if the robot achieves the task in at least 2 out of 3 attempts. Failing one mode results in a success rate of 50\%, while failing both results in 0\%. Success rate is averaged per task.

\textbf{Implementation Details} The robot knowledge graph is built on top of RDFlib. Querying of knowledge graphs is implemented with sparql, while the behavior trees are implemented with py\_trees and ROS. For real-world robot configuration, we use an eye-in-hand Intel RealSense D405 camera and a Kinova Gen3 7-DOF arm. The saliency-based controller proposed in \cite{jiang2024robot} is used for real-time visual servoing across all tasks. 

\begin{figure*}[ht!]
\centering
\includegraphics[width=2.0\columnwidth]{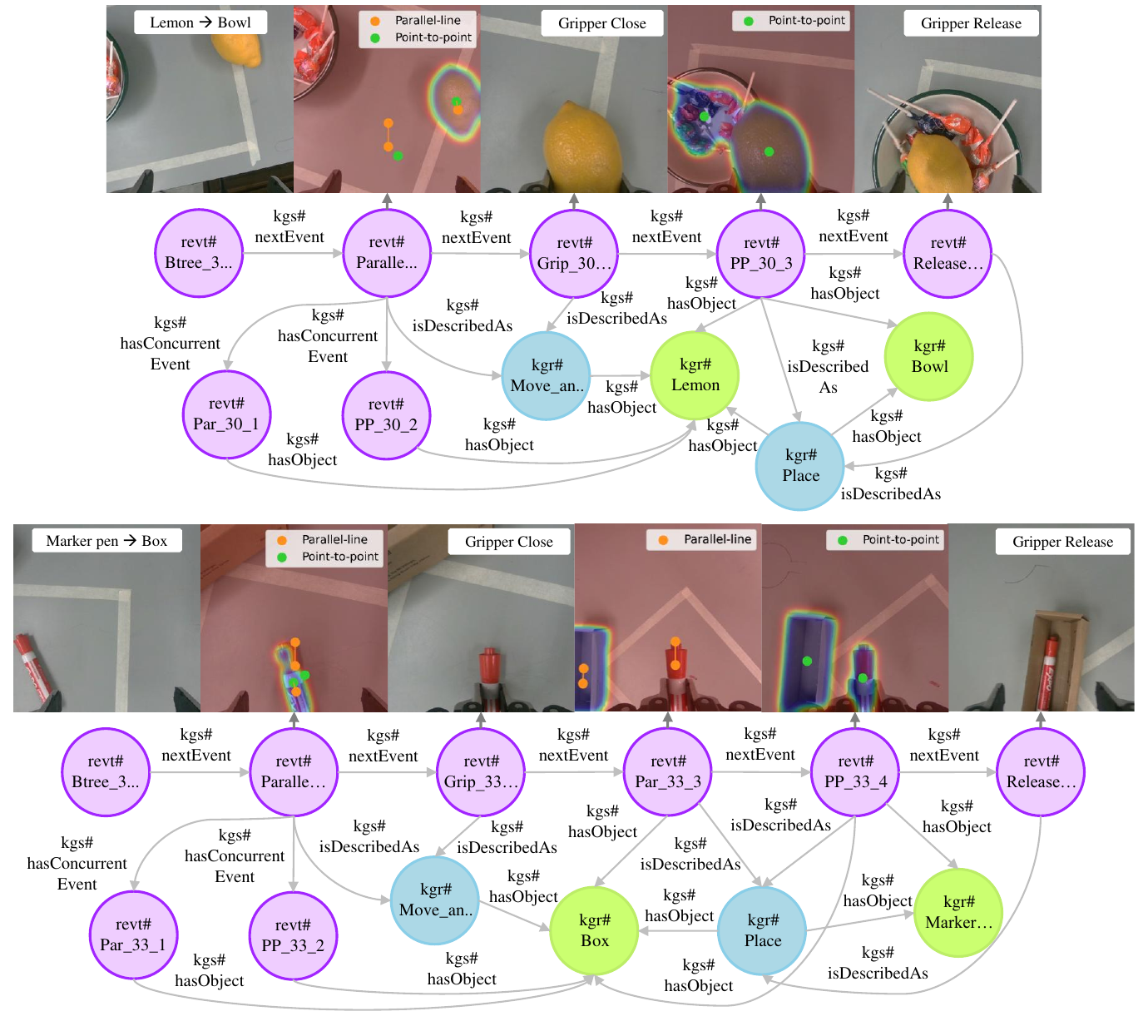}
\caption{Visualization of robot knowledge graphs to conceptualize pick-and-place tasks: (a) lemon $\rightarrow$ bowl; (b) marker pen $\rightarrow$ box. Behaviors and robot events are marked in purple, contexts in blue, and linked entities to DBpedia in green. }
\label{fig:kg}
\end{figure*}

\subsection{Results on Correctness of Constraints}
We report the recall of three methods for predicting geometric constraints in Table \ref{tab:constraints}. Overall, all methods demonstrate innate performance in predicting constraints from context, confirming the connection between manipulation contexts and image geometry. 

\begin{table}[h!]
\centering
\caption{Recall of methods for determining correctness of constraints.}
\label{tab:constraints}
\begin{tabular}{lccc}
\toprule
 & \multicolumn{3}{c}{Method}\\
\cmidrule(lr){2-4}
Task Context & Symbolic & CLIP & GPT4o \cite{jiang2024robot}\\
\midrule
Move-and-Grasp, Food & 77.8\% & 100\% & 66.7\% \\
Move-and-Grasp, Marker Pen & 100\% & 100\% & 100\% \\
Move-and-Grasp, Utility & 100\% & 100\% & 100\% \\
\bottomrule
\end{tabular}
\end{table}

For the symbolic method, 1 out of 12 predictions fails for the lemon grasping task, as by design, the “lemon” concept is not present in the 19 demonstrations used to construct the knowledge base. Since the symbolic method relies solely on querying existing knowledge, it is especially vulnerable to the absence of novel concepts. This limitation could be mitigated by introducing fuzziness in contextual representation, such as canonicalizing the concept “lemon” using more knowledge sources (e.g. \texttt{dbc:Sour\_fruits}). However, effective entity canonicalization is a challenging problem. In practice, combining the symbolic and CLIP-based method to incorporate both conceptual and image context can achieve more robust results.

\subsection{Results on Concurrent Constraints}
Table \ref{Table:Concurrent} reports the average success rate and execution time of concurrent versus sequential constraint optimization. Ideally, optimizing constraints concurrently yields faster execution speeds, as observed in the marker pen grasping task, where optimizing $\varepsilon_{par}$ and $\varepsilon_{p2p}$ in parallel is on average 42\% faster than optimizing in sequence. However, concurrent optimization can be problematic. For example, in bottle grasping task, the controller failed to optimize $\varepsilon_{p2l}$ and $\varepsilon_{p2p}$ concurrently due to rare numerical instability, resulting in dragging and eventual failure. In such cases, sequential optimization is more reliable at the expense of speed.

\begin{table}[h!]
\caption{Average success rate and execution time for concurrent vs. sequential geometric constraints.}
\label{Table:Concurrent}
\centering
\begin{tabular}{l c c c} 
\toprule
Task Context & Average Success Rate & Time \\ 
\midrule
Sequence, Grasp Marker Pen & 100\% & 24.71s \\
Parallel, Grasp Marker Pen & 100\% & 17.37s \\
\midrule
Sequence, Grasp Bottle & 100\% & 21.61s \\
Parallel, Grasp Bottle & 40\% & 23.1s \\

\bottomrule
\end{tabular}
\end{table}

\subsection{Results on Real-world Robot Control}
Table \ref{Table:task} presents the success rates of real-world robot control across the two evaluated task categories. Overall, robot knowledge graphs retain the flexibility of behavior trees for scripting manipulation control, while effectively linking concepts and events at a detailed level. Additionally, Figure \ref{fig:kg} visualizes the executions of two tasks, alongside the corresponding knowledge graphs.

For pick-and-place tasks, constraint specifications are flexible.  For example, for pen-to-box placement, $\varepsilon_{par}$ and $\varepsilon_{p2p}$ can be optimized sequentially or concurrently to achieve the same placement context, which accommodates the perpendicular orientation of the rectangle box. Moreover, the optimizations perform well in both waypoint and velocity control, where the behavior trees successfully handle constraints as blocking or non-blocking actions respectively.

For composition tasks, repeatedly combining constraints gradually builds up the intended task goal, achieving a compositional structure with various contexts over time. Additionally, the flexibility of behavior trees allows integration with other robot behaviors. For example, in the Composition-2 task, the robot must first open the drawer to reveal the tennis ball before grasping it. Scripting such a sequence of contexts requires detailed task conceptualization, which knowledge graphs effectively represent. However, the stability of the vision model is also a crucial factor. In particular, we observe the failure of velocity control in the Composition-1 task, where obscuration over parts of the objects poses significant challenges.

\begin{table}[h!]
\caption{Results with control.}
\label{Table:task}
\centering
\begin{tabular}{c p{4cm} c} 
\toprule
Task & Context & Success Rate \\ 
\midrule
Pick-and-Place & Lemon $\rightarrow$ Bowl. & 100\% \\
 & Apple $\rightarrow$ Basket. & 100\% \\
 & Teabag $\rightarrow$ Mug. & 100\% \\
 & Marker Pen $\rightarrow$ Rectangle Box. & 100\% \\
\midrule
Composition-1 & 2 Fruits \& Candy $\rightarrow$ Basket.  & 50\% \\
Composition-2 & Open Drawer; Grasp Tennis Ball; Tennis Ball $\rightarrow$ Box on table.   & 100\% \\
\bottomrule
\end{tabular}
\end{table}

\section{Conclusions}
We propose to integrate behavior trees and geometric constraints into robot knowledge graphs, enabling the conceptualization and execution of real-world robot manipulation tasks. Through real-world experiments, we demonstrate the effectiveness of the robot knowledge graphs in representing manipulation contexts and events, accomplishing real-world manipulation tasks in daily living scenarios. In future work, we plan to further explore the versatility of robot knowledge graphs by scaling up the variety of daily living tasks. Additionally, automatic generation of robot knowledge graphs presents an opportunity to enhance the autonomy of the robot control, and can be valuable to explore further.







{
\bibliographystyle{IEEEtranS}
\bibliography{citation}
}

\end{document}